\documentclass[preprint,12pt]{elsarticle}

\usepackage{graphicx}
\usepackage{amssymb}
\usepackage{multirow}
\usepackage{tabularx}

\journal{Journal Name}

\begin{document}

\begin{frontmatter}


\title{Textual Data Mining for Financial Fraud Detection: A Deep Learning Approach}



\author{Qiuru Li}

\address{Department of Information Systems, HKUST}

\begin{abstract}
In this report, I present a deep learning approach to conduct a natural language processing (hereafter NLP) binary classification task for analyzing financial-fraud texts. First, I searched for regulatory announcements and enforcement bulletins from HKEX news to define fraudulent companies and to extract their MD\&A reports before I organized the sentences from the reports with labels and reporting time. My methodology involved different kinds of neural network models, including Multilayer Perceptrons with Embedding layers, vanilla Recurrent Neural Network (RNN), Long-Short Term Memory (LSTM), and Gated Recurrent Unit (GRU) for the text classification task. By utilizing this diverse set of models, I aim to perform a comprehensive comparison of their accuracy in detecting financial fraud. My results bring significant implications for financial fraud detection as this work contributes to the growing body of research at the intersection of deep learning, NLP, and finance, providing valuable insights for industry practitioners, regulators, and researchers in the pursuit of more robust and effective fraud detection methodologies.
\end{abstract}

\end{frontmatter}


\section{Introduction}
\label{S:1}

NLP, as a key node of Artificial Intelligence, has been developed as one of the trendiest techniques in not only the finance and accounting literature for financial texts analysis \cite{fisher2016natural, huang2023finbert}, but also in the financial practice, for instance, risk assessments, auditing and stock behavior predictions \cite{kang2020natural}; Healthcare and automotive industries are the leaders of NLP nowadays. However, there are limitations for NLP algorithms given that they can neither understand the entire context and analyze a text as a bag of individual words nor learn the grammar and word order \cite{loughran2016textual}. On such grounds, deep learning NLP were evolved to diminish these insufficiencies, for example, word2vec \cite{mikolov2013efficient}, RNN \cite{rumelhart1985learning}, RNN variants (LSTM \cite{hochreiter1997long}, GRU \cite{chung2014empirical}, etc.), and Attention Mechanism \cite{vaswani2017attention} which can conduct various text analysis tasks, including sentiment analysis, sentence-level classification and image captioning. With deep learning NLP algorithms, we can dig out more insights from the analysis of human languages.

RNN, with the recurring mechanism which can feed the results back to the neural network, is powerful for analyzing sequential data, such as texts and videos. Though RNN considers the previous and current information at the same time, it has a short-term memory problem due to the vanishing gradient problem. Therefore, RNN variants, GRU and LSTM were innovated to tackle these deficiencies of RNN. LSTM and GRU are the most widely used RNN variants as they can easily learn long-distance dependencies. Besides, there are bidirectional RNNs and multi-layer RNNs which have higher computation capability for more complicated representations further improving the performance of RNNs. Therefore, in this paper, I am going to learn the performance of different RNN algorithms through the task of financial fraud detection, classifying fraudulent and non-fraudulent sentences, and comparing the accuracy of various RNN models. With this research, regulatory authorities and financial companies can gain more useful insights into developing financial fraud detection techniques with the application of deep learning NLP algorithms.

\section{Methodology}
\label{s:2}

\subsection{Deep Learning NLP Algorithms}

\subsubsection{Recurrent Neural Networks (RNN)}

RNN is a type of neural network where the output from the previous step is fed as the input to the current step. RNN has the present and the recent past as two inputs and applies weights to these inputs. The weights are adjusted for both backpropagations through time and gradient descent. Within backpropagation through time, the error is backpropagated from the last to the first timestep, while unrolling all the timesteps. This allows calculating the error for each timestep, which allows updating the weights. The gradient is for updating the weights and exponentially shrinks up or down during the backpropagation, which can lead to exploding gradient or vanishing gradient that makes RNN difficult to retain information in the earlier layers.

With internal memory, RNN is able to handle sequential or time series data and temporal problems, such as language translation and natural language processing. RNN is used by Apple’s Siri and Google’s voice search.

\subsubsection{Long-Short Term Memory (LSTM)}

LSTM is an advanced RNN that can also learn the dependence of sequential data, which mitigates the vanishing gradient descent issue by preserving the information from earlier time steps. LSTM is powerful in speech recognition, video analysis, and sentiment analysis. Moreover, LSTM can handle complicated datasets accurately.
 
LSTM has three gates: forget gate decides what is kept and forgotten from the previous cell state; The input gate controls what parts of the new cell content are written to the cell; The output gate determines what parts of the cell are output to hidden state. The picture below illustrates the operations of LSTM.

\begin{figure}[h!]
\centering\includegraphics[width=0.8\linewidth]{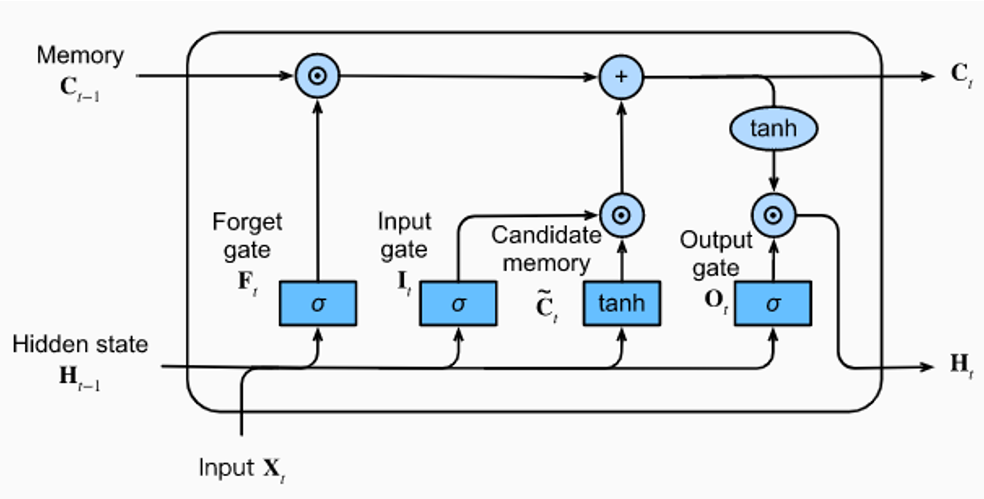}
\caption{LSTM's gate structures.}
\end{figure}

\subsubsection{Gated Recurrent Unit (GRU)}

GRU is an advanced version of RNN that can also tackle sequential data and resolve the problem of vanishing gradient descent. GRU is very similar to LSTM, but with only two gates, which makes it faster in training and more preferred to small datasets since GRU takes fewer parameters.

GRU has a reset gate, which is responsible for short-term memory and controls what parts of the hidden state are updated or preserved, and an update gate for long-term memory, which decides what parts of the previous hidden state are used to compute new content separately. The picture below shows the operations of GRU.

\begin{figure}[h!]
\centering\includegraphics[width=0.8\linewidth]{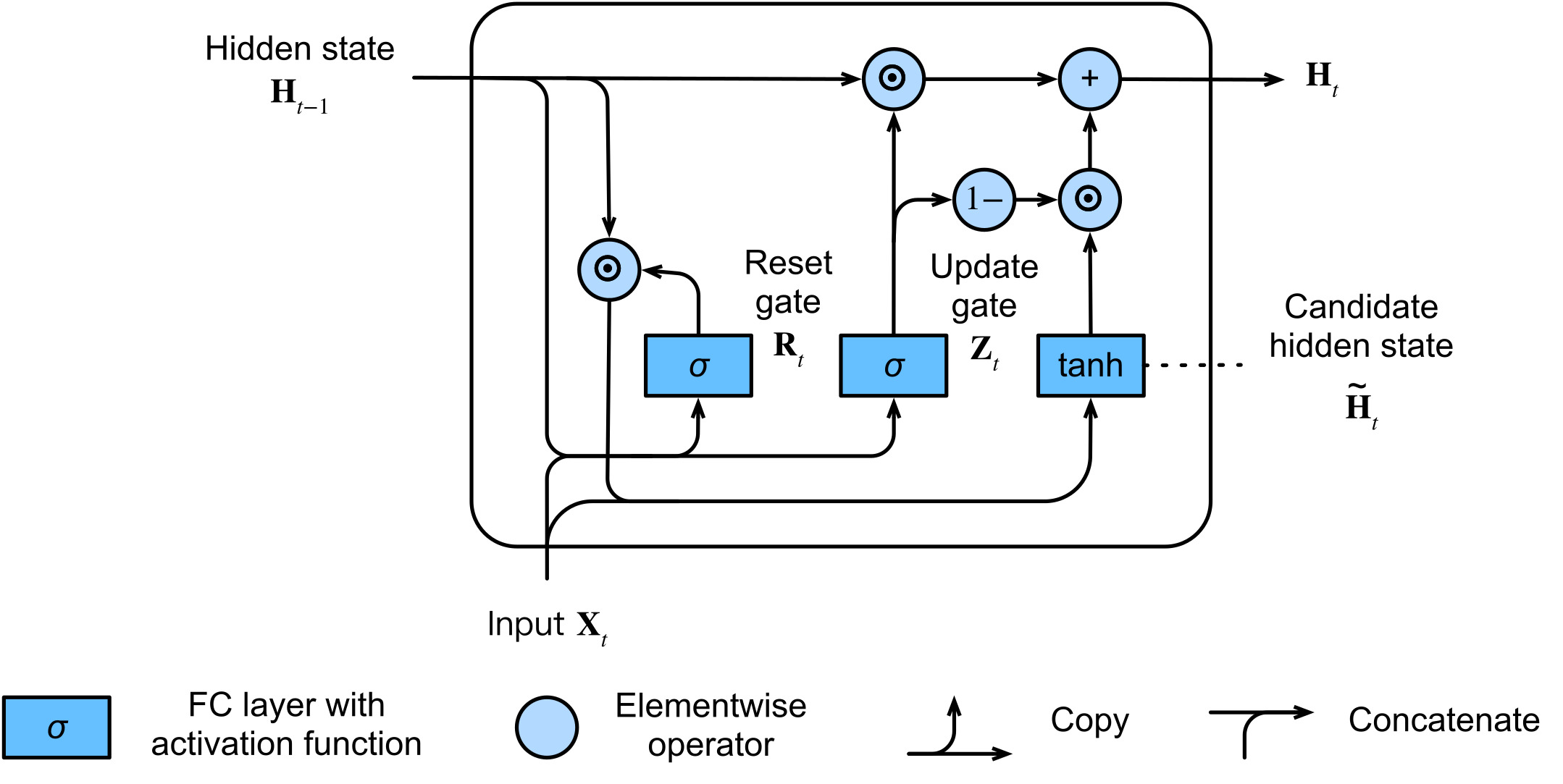}
\caption{LSTM's gate structures.}
\end{figure}

\subsubsection{Bidirectional recurrent neural networks (BRNN)}

BRNN connects two hidden layers of opposite directions to the same output. BRNN can be trained without the limitation of using input information just up to a present future frame. This is accomplished by training it simultaneously in the positive and negative time directions \cite{650093}, hence, the output layer can get information from backward and forward states at the same time. BRNN is powerful if there is access to the entire input sequence, and especially useful when the context of the input is needed.

\subsection{Data Collection and Organization}

In order to conduct financial fraud detection, we have searched for fraudulent cases reported by HKEX news from December 2000 to July 2022. Then we defined the case period, fraudulent theme, relevant document, and reported time through reading the news. According to this organized information, we downloaded relevant documents, i.e., financial statement, annual report, and MD\&A, for the case period and extracted sentences from the documents. Finally, we labeled the sentences and time of the case in order to conduct further model training. The label includes “F” as fraudulent and “NF” as non-fraudulent.

\subsection{Data processing}

We have 34838 samples with features “textual data”, “label” and “time” in total. First, to make it more convenient to group and classify, I converted the label into a numeric variable. “F” is represented by 1 and “NF” is represented by 0. Second, I conducted data cleansing by removing stop words for textual data columns and split the whole dataset into train and test sets. Then, I further cleaned and organized the two datasets by separating textual data and labels, removing symbols, and lowering the case for textual data. From the collected data I observed that the number of “F” and “NF” labeled sentences are very imbalanced, I used oversampling to increase the sample number of the minority class. Finally, I applied a tokenizer to generate the word index dictionary and list of token sequences from the training texts, then pad the sequences to a uniform length for both training and test texts to prepare for the model training and fitting.

\subsection{Model building}

Considering the training parameters, I use the binary cross entropy loss since the financial-fraud text detection is a binary classification task; Adam as the optimizer because it has faster convergence and can easily handle heavy-tailed noise; AUC (Area Under Curve) as the evaluation metric since our dataset is imbalanced, using AUC can avoid fake high accuracy since AUC reflects true positive rate and false positive rate of prediction.
For the model fitting parameters, since the dataset is large and imbalance, I use a large batch size and train for a small number of epochs to enhance the predicting efficiency.

\begin{table}[]
\begin{tabularx}{1.1\textwidth}{|X|X|X|X|}
\hline\vspace{5pt}
\hspace{25pt}
 Model & \vspace{5pt}\hspace{3pt}Hyperparameters & \vspace{5pt}\hspace{30pt}Layers & \vspace{1pt}Training parameters \\ [0.5ex] 
 \hline\hline
 \vspace{5pt}Simple NN with Embedding layers &  & 1 Embedding layer, 1 flatten layer and 2 dense layers &  \\ 
 \cline{1-1}  \cline{3-3}
 \vspace{20pt}\hspace{15pt}Vanilla RNN & & 1 Embedding layer, 1 bidirectional vanilla RNN layer and 2 dense layers &  \\
  \cline{1-1}\cline{3-3}
 \vspace{40pt}\hspace{30pt}LSTM & \vspace{30pt}vocab\_size = 20000, embedding\_dim = 150, input\_length = maxlen = 200 & 1 Embedding layer, 1 Bidirectional LSTM layer, 2 dense layers, 2 dropout layers and 1 Batch Normalization layer & \vspace{30pt}Loss: binary cross entropy, Optimizer: Adam, Metrics: AUC, batch\_size = 256, epoch = 10 \\
  \cline{1-1}\cline{3-3}
 \vspace{15pt}\hspace{33pt}GRU &  & 1 Embedding layer, 1 bidirectional GRU layer, 2 dense layers &  \\
  \cline{1-1}\cline{3-3}
 \vspace{40pt}Multi-layer LSTM &  & 1 Embedding layer, 2 Bidirectional LSTM layers, 2 dense layers, 2 dropout layers and 1 Batch Normalization layer &  \\ [1ex] 
 \hline

\end{tabularx}
 \caption{An overview of model architecture and training configurations.}
\end{table}

\section{Results}
\label{s:3}
To illustrate easier comparison among neural network models, I extracted the performance data from the epoch with best validation accuracy for reference.

\begin{table}[h!]
\centering

 \begin{tabularx}{\textwidth}{|X|X|X|X|X|}
 \hline
 \vspace{0.1pt}Model & \vspace{0.1pt}Training loss & \vspace{0.1pt}Training AUC & \vspace{0.1pt}Validation loss & \vspace{0.1pt}Validation AUC \\ [0.5ex] 
 \hline\hline
 Simple NN with embedding layers & 0.0350 & 0.9998 & 0.0612 & 0.5112 \\ 
 \hline
 Vanilla RNN & 0.0006 & 1.0000 & 0.0564 & 0.5118 \\
 \hline
 LSTM & 0.0014 & 1.0000 & 0.0545 & 0.5087 \\
 \hline
 GRU & 0.0002 & 1.0000 & 0.0991 & 0.5111 \\
 \hline
 Multi-layer LSTM & 0.0130 & 0.9999 & 0.6405 & 0.5990\\ [1ex] 
 \hline
 \end{tabularx}
 \caption{Training and validation statistics of different models.}
 
\end{table}

\section{Discussion}
\label{s:4}
As the training loss and AUC describes how well the model performs on the training dataset and the validation loss and AUC implies how well the model fits the new data, the model should be regarded as well-performed if it has a higher AUC for both the training and validation dataset. The results of the best model regarding the validation AUC reveal that the multi-layer LSTM model is slightly better than the other models, and the overall validation loss shows that the LSTM is more well-performed since the LSTM model’s validation loss are decreasing during the training and predicting while the others are increasing. However, the overall performance shows that the validation AUC for the final model is around 0.5 and the ultimate training AUC achieved 1 for all five models, which reflects the overfitting problem and the low prediction abilities of the five neural models. However, in principle, neural network models, especially RNNs should perform well in tackling text classification tasks. From my point of view, the reason of unsatisfactory prediction results in this
 
financial-fraud text detection task can be the imbalance of our samples. The number of samples with the label “NF” is 175 times of samples with the label “F”, which can lead to bias towards the majority class only. Additionally, in real life, a majority of companies won’t be fraudulent.
Therefore, it will be difficult for the models to predict the minority fraudulent class. Although the application of deep learning NLP algorithms is powerful and practical in the financial industry currently, it is challenging to conduct financial-fraud text detection.

\section{Acknowledgments}
\label{s:5}
I would like to express my sincere gratitude to the organizers of the Undergraduate Research Opportunity Program (UROP), which provides a platform for students to conduct research. Moreover, I would like to show my appreciation to Professor HUANG, and Professor Yang since they gave me great support throughout the whole research project, and also my research teammates as we collaborated together to tackle the obstacles.





\bibliographystyle{model1-num-names}
\bibliography{sample.bib}

\begin{thebibliography}{10}
\expandafter\ifx\csname natexlab\endcsname\relax\def\natexlab#1{#1}\fi
\providecommand{\bibinfo}[2]{#2}
\ifx\xfnm\relax \def\xfnm[#1]{\unskip,\space#1}\fi
\bibitem[{Fisher et~al.(2016)Fisher, Garnsey, and Hughes}]{fisher2016natural}
\bibinfo{author}{I.~E. Fisher}, \bibinfo{author}{M.~R. Garnsey},
  \bibinfo{author}{M.~E. Hughes},
\newblock \bibinfo{title}{Natural language processing in accounting, auditing
  and finance: A synthesis of the literature with a roadmap for future
  research},
\newblock \bibinfo{journal}{Intelligent Systems in Accounting, Finance and
  Management} \bibinfo{volume}{23} (\bibinfo{year}{2016})
  \bibinfo{pages}{157--214}.
\bibitem[{Huang et~al.(2023)Huang, Wang, and Yang}]{huang2023finbert}
\bibinfo{author}{A.~H. Huang}, \bibinfo{author}{H.~Wang},
  \bibinfo{author}{Y.~Yang},
\newblock \bibinfo{title}{Finbert: A large language model for extracting
  information from financial text},
\newblock \bibinfo{journal}{Contemporary Accounting Research}
  \bibinfo{volume}{40} (\bibinfo{year}{2023}) \bibinfo{pages}{806--841}.
\bibitem[{Kang et~al.(2020)Kang, Cai, Tan, Huang, and Liu}]{kang2020natural}
\bibinfo{author}{Y.~Kang}, \bibinfo{author}{Z.~Cai}, \bibinfo{author}{C.-W.
  Tan}, \bibinfo{author}{Q.~Huang}, \bibinfo{author}{H.~Liu},
\newblock \bibinfo{title}{Natural language processing (nlp) in management
  research: A literature review},
\newblock \bibinfo{journal}{Journal of Management Analytics}
  \bibinfo{volume}{7} (\bibinfo{year}{2020}) \bibinfo{pages}{139--172}.
\bibitem[{Loughran and McDonald(2016)}]{loughran2016textual}
\bibinfo{author}{T.~Loughran}, \bibinfo{author}{B.~McDonald},
\newblock \bibinfo{title}{Textual analysis in accounting and finance: A
  survey},
\newblock \bibinfo{journal}{Journal of Accounting Research}
  \bibinfo{volume}{54} (\bibinfo{year}{2016}) \bibinfo{pages}{1187--1230}.
\bibitem[{Mikolov et~al.(2013)Mikolov, Chen, Corrado, and
  Dean}]{mikolov2013efficient}
\bibinfo{author}{T.~Mikolov}, \bibinfo{author}{K.~Chen},
  \bibinfo{author}{G.~Corrado}, \bibinfo{author}{J.~Dean},
\newblock \bibinfo{title}{Efficient estimation of word representations in
  vector space},
\newblock \bibinfo{journal}{arXiv preprint arXiv:1301.3781}
  (\bibinfo{year}{2013}).
\bibitem[{Rumelhart et~al.(1985)Rumelhart, Hinton, Williams
  et~al.}]{rumelhart1985learning}
\bibinfo{author}{D.~E. Rumelhart}, \bibinfo{author}{G.~E. Hinton},
  \bibinfo{author}{R.~J. Williams}, et~al., \bibinfo{title}{Learning internal
  representations by error propagation}, \bibinfo{year}{1985}.
\bibitem[{Hochreiter and Schmidhuber(1997)}]{hochreiter1997long}
\bibinfo{author}{S.~Hochreiter}, \bibinfo{author}{J.~Schmidhuber},
\newblock \bibinfo{title}{Long short-term memory},
\newblock \bibinfo{journal}{Neural computation} \bibinfo{volume}{9}
  (\bibinfo{year}{1997}) \bibinfo{pages}{1735--1780}.
\bibitem[{Chung et~al.(2014)Chung, Gulcehre, Cho, and
  Bengio}]{chung2014empirical}
\bibinfo{author}{J.~Chung}, \bibinfo{author}{C.~Gulcehre},
  \bibinfo{author}{K.~Cho}, \bibinfo{author}{Y.~Bengio},
\newblock \bibinfo{title}{Empirical evaluation of gated recurrent neural
  networks on sequence modeling},
\newblock \bibinfo{journal}{arXiv preprint arXiv:1412.3555}
  (\bibinfo{year}{2014}).
\bibitem[{Vaswani et~al.(2017)Vaswani, Shazeer, Parmar, Uszkoreit, Jones,
  Gomez, Kaiser, and Polosukhin}]{vaswani2017attention}
\bibinfo{author}{A.~Vaswani}, \bibinfo{author}{N.~Shazeer},
  \bibinfo{author}{N.~Parmar}, \bibinfo{author}{J.~Uszkoreit},
  \bibinfo{author}{L.~Jones}, \bibinfo{author}{A.~N. Gomez},
  \bibinfo{author}{{\L}.~Kaiser}, \bibinfo{author}{I.~Polosukhin},
\newblock \bibinfo{title}{Attention is all you need},
\newblock \bibinfo{journal}{Advances in neural information processing systems}
  \bibinfo{volume}{30} (\bibinfo{year}{2017}).
\bibitem[{Schuster and Paliwal(1997)}]{650093}
\bibinfo{author}{M.~Schuster}, \bibinfo{author}{K.~Paliwal},
\newblock \bibinfo{title}{Bidirectional recurrent neural networks},
\newblock \bibinfo{journal}{IEEE Transactions on Signal Processing}
  \bibinfo{volume}{45} (\bibinfo{year}{1997}) \bibinfo{pages}{2673--2681}.

\end{thebibliography}





\end{document}